\documentclass{article}

\usepackage[nonatbib, final]{nips_2017}

\usepackage[utf8]{inputenc} 
\usepackage[T1]{fontenc}    
\usepackage{hyperref}       
\usepackage{url}            
\usepackage{booktabs}       
\usepackage{amsfonts}       
\usepackage{nicefrac}       
\usepackage{microtype}      
\usepackage{graphicx}

\usepackage{algorithm}
\usepackage[noend]{algpseudocode}

\title{Federated Control with Hierarchical Multi-Agent Deep Reinforcement Learning}

\author{
  Saurabh Kumar\thanks{Equal contribution. Work done while Saurabh Kumar interned at Google Research.} \\
  Georgia Tech \\
  \texttt{skumar311@gatech.edu} \\
  \And
  Pararth Shah\footnotemark[1] \\
  Google \\
  \texttt{pararth@google.com} \\
  \AND
  Dilek Hakkani-T\"ur \\
  Google \\
  \texttt{dilekh@ieee.org} \\
  \And
  Larry Heck \\
  Google \\
  \texttt{larry.heck@ieee.org} \\
}

\begin{document}

\maketitle

\begin{abstract}
We present a framework combining hierarchical and multi-agent deep reinforcement learning approaches to solve coordination problems among a multitude of agents using a semi-decentralized model. The framework extends the multi-agent learning setup by introducing a meta-controller that guides the communication between agent pairs, enabling agents to focus on communicating with only one other agent at any step. This hierarchical decomposition of the task allows for efficient exploration to learn policies that identify globally optimal solutions even as the number of collaborating agents increases. We show promising initial experimental results on a simulated distributed scheduling problem.
\end{abstract}

\section{Introduction}
Multi-agent reinforcement learning \cite{foerster2017counterfactual} can be applied to many real-world coordination problems, e.g. network packet routing or urban traffic control, to find decentralized policies that jointly optimize the private value functions of participating agents. However, multi-agent RL algorithms scale poorly with problem size. Since communication possibilities increase quadratically as the number of agents, the agents must explore a larger combined action space before receiving feedback from the environment. In a separate line of research, hierarchical reinforcement learning (HRL) \cite{kulkarni2016hierarchical} has enabled learning goal-directed behavior from sparse feedback in complex environments. HRL divides the overall task into independent goals and trains a meta-controller to pick the next goal, while a controller learns to reach individual goals. Consequently, HRL requires the task to be divisible into independent subtasks that can be solved sequentially. Multi-agent problems do not directly fit this criteria as a subtask would entail coordination between multiple agents, each having partial observability on the global state. Recently, \cite{lewis2017deal} trained a pair of agents to negotiate and agree upon a joint action, but they do not explore scaling to settings with many agents.

We propose Federated Control with Reinforcement Learning (FCRL), a framework for combining hierarchical and multi-agent deep RL to solve multi-agent coordination problems with a semi-decentralized model. Similar to HRL, the model consists of a \textit{meta-controller} and \textit{controllers}, which are hierarchically organized deep reinforcement learning modules that operate at separate time scales. In contrast to HRL, we modify the notion of a controller to characterize a decentralized agent which receives a partial view of the state and chooses actions that maximize its private value function. The model supports a variable number of controllers, where each controller is intrinsically motivated to negotiate with another controller and agree upon a joint action under some constraints, e.g. a division of available resources or a consistent schedule. The meta-controller chooses a sequence of pairs of controllers that must negotiate with each other as well as a constraint provided to each pair, and it is rewarded by the environment for efficiently surfacing a globally consistent set of controller actions. Since a controller needs to communicate with a single other controller at any step, the controller's policy can be trained separately via self-play to maximize expected future intrinsic reward with gradient descent. As the details of individual negotiations are abstracted away from the meta-controller, it can efficiently explore the space of choices of controller pairs even as number of controllers increases, and it is trained to maximize expected future extrinsic reward with gradient descent.

FCRL can be applied to a variety of real-world coordination problems where privacy of agents' data is paramount. An example is multi-task dialogue with an automated assistant, where the assistant must help a user to complete multiple interdependent tasks, for example making a plan to take a train to the city, watch a movie and then get dinner. Each task requires querying a separate third-party Web service which has a private database of availabilities, e.g. a train ticket purchase, a movie ticket purchase or a restaurant table reservation service. Each Web service is a decentralized controller which aims to maximize its utilization, while the assistant is a meta-controller which aims to obtain a globally viable schedule for the user. 
Another example is urban traffic control, where each vehicle is a controller having a destination location that it desires to keep private. A meta-controller guides the traffic flow through a grid of intersections, aiming to maintain a normal level of traffic on all roads in the grid. The meta-controller iteratively picks a pair of controllers and road segments, and the controllers must negotiate with each other and assign different road segments among themselves.

In the next section, we formally describe the Federated RL model, and in Section 3 we mention related work. In Section 4 we present preliminary experiments with a simulated multi-task dialogue problem. We conclude with a discussion and present directions for future work in Section 5.

\section{Model}
\begin{figure}[t]
  \center
  \includegraphics[width=0.48\linewidth]{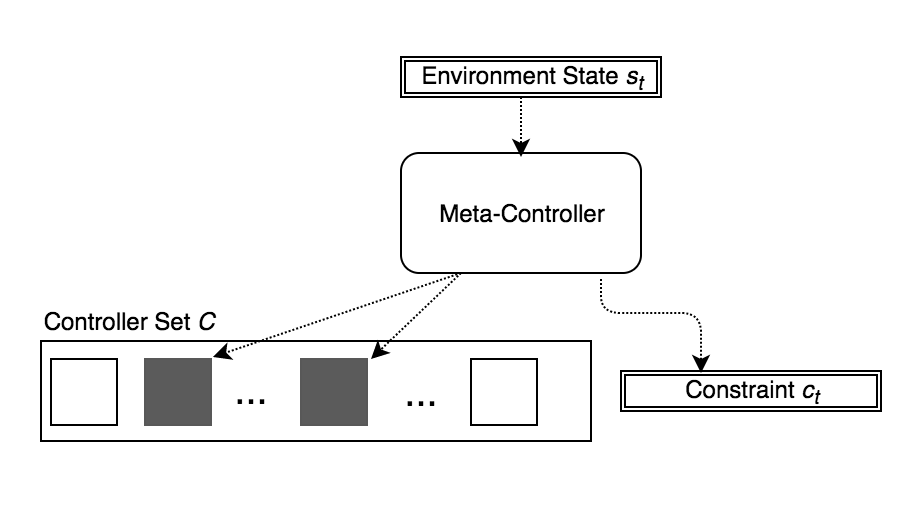}
  \includegraphics[width=0.48\linewidth]{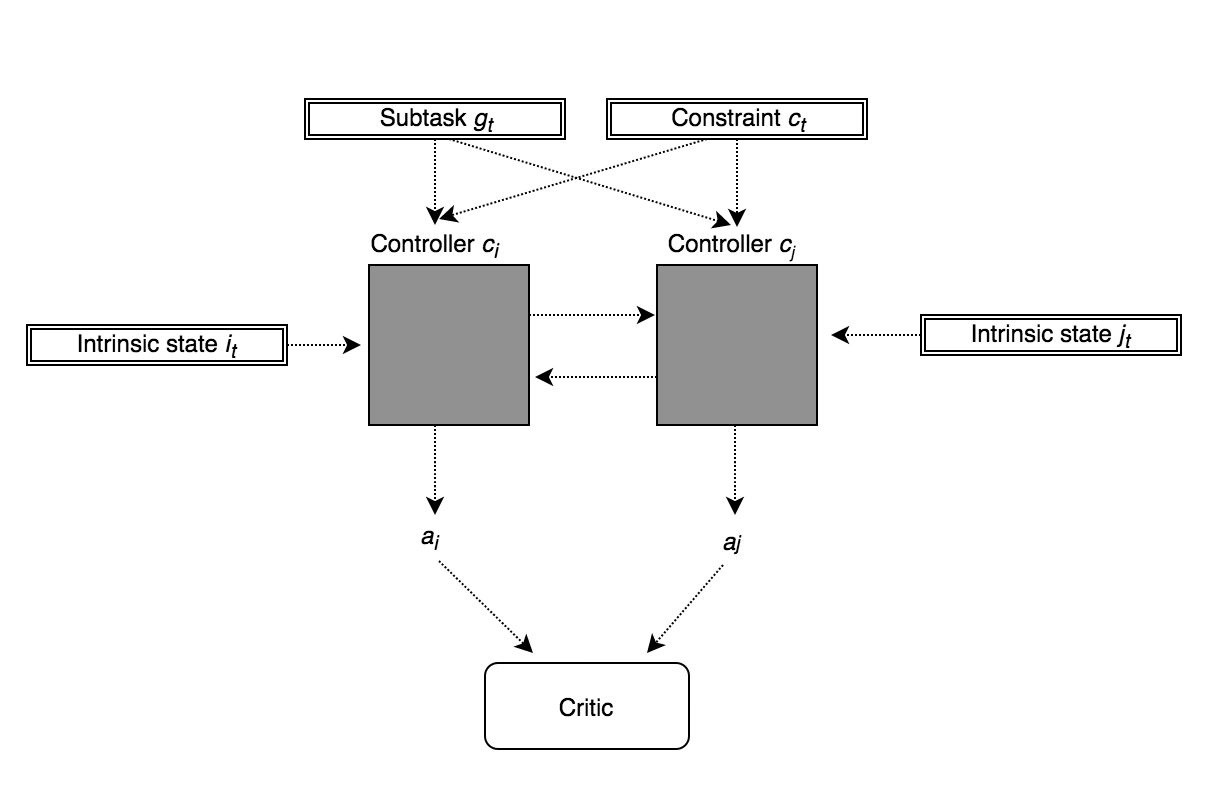}
  \caption{Federated control model}
  \label{fig:tagger}
\end{figure}

Reinforcement Learning (RL) problems are characterized by an agent interacting with a dynamic environment with the objective of maximizing a long term reward \cite{sutton1998reinforcement}. The basic RL model casts the task as a Markov Decision Process (MDP) defined by the tuple $\{S, A, T, R, \gamma\}$ of states, actions, transition function, reward, and discount factor.

\textbf{Agents} As in the h-DQN setting \cite{kulkarni2016hierarchical}, we construct a two-stage hierarchy with a meta-controller and a controller that operate at different temporal scales. However, rather than utilizing a single controller which learns to complete multiple subtasks, FCRL employs multiple controllers, each of which learns to communicate with another controller to collectively complete a prescribed subtask.

\textbf{Temporal Abstractions} As shown in Figure 1, the meta-controller receives a state $s_t$ from the environment and selects a subtask $g_t$ from the set of all subtasks and a constraint $c_t$ from the set of all constraints. Constraints ensure that individual subtasks focus on disjoint parts of the overall problem, allowing each problem to be solved independently by a subset of the controllers. The meta-controller’s goal is to pick a sequence of subtasks and associated constraints to maximize the cumulative discounted extrinsic reward provided by the environment. A subtask $g_t$ is associated with two controllers, $C_i$ and $C_j$, who must communicate with each other to complete the task. $C_i$ and $C_j$ receive separate partial views of the environment state through states $i_t$ and $j_t$. For $K-1$ time steps, the subtask $g_t$ and constraint $c_t$ remain fixed while $C_i$ and $C_j$ negotiate to decide on a set of output actions, $a_i$ and $a_j$, which are outputted at the $K$th time step. The controllers are trained with an intrinsic reward provided by an internal critic. The reward is shared between the controller pairs that communicated with each other, and they are rewarded for choosing actions that satisfy the constraints provided by the environment as well as the meta-controller.

\section{Related Work}

\textbf{Multi-agent communication with Deep RL} Multi-agent RL involves multiple agents that must either cooperate or compete in order to successfully complete a task. A number of recent works have demonstrated the success of this approach and have applied it to tasks in which agents communicate with one another \cite{foerster2016learntocomm, mordatch2017complang, sukhbaatar2016learning}. Two agents learned to communicate in natural language with one another in order to complete a negotiation task in \cite{lewis2017deal} and an image guessing task in \cite{das2017visdial}. The demonstrated success of multi-agent communication is promising, but it may be difficult to scale to greater numbers of agents. In our work, we combine multi-agent RL with a meta-controller that selects subsets of agents to communicate so that the overall task is optimally completed.

\textbf{Hierarchical Deep RL} The h-DQN algorithm \cite{kulkarni2016hierarchical} splits an agent into two components: a meta-controller that selects subtasks to complete and a controller that selects primitive actions given a subtask as input from the meta-controller. FCRL uses multiple controllers and extends the notion of a subtask to involve a subset of the controllers that must collectively communicate to satisfy certain constraints. Therefore, each controller can be pre-trained to complete a distinct goal and may receive information only relevant for that particular goal. This is in contrast to the work by \cite{kulkarni2016hierarchical}, which trains a single controller to complete multiple subtasks.

\textbf{Hierarchical Deep RL for dialogue} In task-oriented dialogues, a dialogue agent assists a user in completing a particular task, such as booking movie tickets or making a restaurant reservation. Composite tasks are those in which multiple tasks must be completed by the dialogue agent. Recent work by \cite{peng2017composite} applies the h-DQN technique to composite task completion. While this work successfully trains agents on composite task dialogues, a drawback with the straightforward application of h-DQN to dialogue is that only one goal is in focus at any given time, which must be completed prior to another goal being addressed. This prevents the dialogue agent from handling cross-goal constraints. The FCRL algorithm addresses cross-goal constraints by modeling a subtask as a negotiation between two controllers that must simultaneously complete their individual goals.

\section{Experiments}
We present a preliminary experiment applying our method to a simulated distributed scheduling problem. The goal is to validate our approach against baseline approaches in a controlled setup. We plan to run further experiments on realistic scenarios in future work.

\subsection{Environment}
We consider a distributed scheduling problem which is inspired by the multi-domain dialogue management setup described in the introduction. Formally, this consists of $N$ agents, each having a private database of available time entries, who must each pick a time such that the relative order of times chosen by all agents is consistent with the order specified by the environment. At the start of an episode, the environment randomly chooses $m$ agents and provides an ordering\footnote{We model this directly as a sequence of agent IDs, but an extension is to generate or sample crowd-sourced utterances for the constraints (``I want to watch a movie and get dinner. Also I'll need a cab to get there.'') and train the agents to parse the natural language into an agent order.} of agents $C_1, C_2, \dots, C_m$ and agent databases $D_1, \dots, D_m$, where $D_i$ is a bit vector of size $B$ specifying the times that are available to agent $i$. The agents can communicate amongst each other for $K-1$ rounds, after which each agent must output an action $0 \leq a_i < B$. The environment emits a reward $R=1$ if the actions $a_1, \dots a_m$ are such that $a_1 < a_2 < \dots < a_m$, and $a_i \in D_i$ $\forall i$, else it emits a reward of $R=0$. In our setup we used $N=20$, $B=8$, and experimented with $m \in \{2, 4, 6\}$.

\subsection{Agents}
We evaluate three approaches for solving the distributed scheduling problem. Our proposed approach (FCRL) consists of a meta-controller that picks a pair of controllers and a constraint vector, and controller agents which communicate in pairs and output times that satisfy their private databases as well as the meta-controller's constraint. The two baselines, Multi-agent RL (MARL) and Hierarchical RL (HRL), compare our approach with the settings without a meta-controller or multi-agent communication, respectively. 

\textbf{Federated Control (FCRL)} We use an FCRL agent with $K=2$ communication steps between the controllers. Note that this means they communicate for $K-1$ steps and then produce the output actions at the $K$th time step. The controller and meta-controller Q-networks have the same structure with two hidden layers of sizes $100, 50$, each followed by a $tanh$ nonlinearity, and a final fully-connected layer outputting Q-values for each action. The controller has $B$ actions, one for each possible time value, and the meta-controller has $B-1$ actions, corresponding to constraint windows of sizes $B/2^{j}$, $j \in [0, log B]$. (We assume that $B$ is a power of 2.) The meta-controller iterates through agent pairs in the order expected by the environment, and for the pair selected at time t, it chooses a constraint vector $c_t$. This constraint is applied to the two agents' databases, and the controllers then communicate with each other for $K=2$ rounds. If the controllers are able to come up with a valid order, they are rewarded by the intrinsic critic, and the meta-controller moves on to the next pair. Otherwise, the meta-controller retries the same pair of controllers. The meta-controller is given a maximum of 10 total invocations of controller pairs, after which the episode is terminated with reward $R=0$.

For a controller $C_i$ communicating with another controller $C_j$, $C_i$'s state is a concatenation of the database vector $D_i$ from the environment, a one-hot vector of size $2$ denoting the position of that agent in the relative order between $C_i$ and $C_j$, and a communication vector of size $B$ which is a one-hot vector denoting $C_j$'s output in the previous round. (In the first round the communication vector is zeroed out.) The meta-controller's  state is a concatenation of a one-hot vector of size $B$ denoting the latest time entry that has been selected so far, and a multi-hot vector of size $B-1$, denoting the constraints that have been tried for the current controller pair.

\textbf{Multi-agent RL (MARL)}
As a baseline, we consider a setup without a meta-controller, which is the standard multi-agent setup where agents communicate with each other and emit individual actions. The controller agent is same as that in FCRL, except that the position vector is a one-hot vector of size $m$, denoting $C_i$'s position in the overall order, and the communication vector is an average of the outputs of all other agents in the previous round, similar to CommNet described in \cite{sukhbaatar2016learning}.

\textbf{Hierarchical RL (HRL)}
We also consider a Hierarchical RL baseline as described in \cite{kulkarni2016hierarchical}. The controllers do not communicate with each other but instead independently achieve the task of emitting a time value that is consistent with their database and the meta-controller's constraint. The meta-controller is the same as FCRL, except that it picks one agent at a time and assigns it a constraint vector.

\subsection{FCRL Training}
Below, we present the pseudocode for training the FCRL agent.

\begin{algorithm}
\caption{Learning algorithm for FCRL agent}\label{alg:euclid}
\begin{algorithmic}[1]
\State Initialize experience replay buffer $R_M$ for meta-controller and $R_C$ for the controllers 
\State Intialize meta-controller's Q-network, $Q_1$, and controllers' Q-network, $Q_2$, with random weights

\For{$episode$ = 1:N}
\State Environment selects $m$ controllers $C_1$, $C_2$, ..., $C_m$ with databases $D_1$, $D_2$, ..., $D_m$
\State $s \gets \{dp, dt, tc\} \gets \{\phi, \phi\, \phi\}$  \Comment{Meta-controller state is: done pairs ($dp$), done times ($dt$), tried constraints ($tc$)}
\State $t \gets 0$
\While{$s$ is not terminal}
\State $C_i, C_j \gets NextControllerPair(s)$
\State $c_t \gets epsilon\_greedy(\pi(c_t|s), \epsilon_M)$
\State $s_{Ci} \gets \{D_i, c_t, \phi\}$
\State $s_{Cj} \gets \{D_j, c_t, \phi\}$
\For {$communication$ $turn$ = 1:K}
\State $a_i \gets epsilon\_greedy(\pi(a_i|s_{Ci}), \epsilon_{Ci})$
\State $a_j \gets epsilon\_greedy(\pi(a_j|s_{Cj}), \epsilon_{Cj})$
\State $s'_{Ci} \gets \{D_i, c_t, a_j\}$
\State $s'_{Cj} \gets \{D_j, c_t, a_i\}$
\State $r_{intrinsic} \gets Critic(s, a_i, a_j)$
\State Store transition ($s_{Ci}$, $a_i$, $r_{intrinsic}$, $s'_{Ci}$) in $R_C$
\State Store transition ($s_{Cj}$, $a_j$, $r_{intrinsic}$, $s'_{Cj}$) in $R_C$

\State $s_{Ci} \gets s'_{Ci}$
\State $s_{Cj} \gets s'_{Cj}$

\EndFor

\State Sample minibatch of transitions from $R_C$ and update $Q_2$ weights 
\If{$r_{intrinsic} > 0$} \Comment{Controller pair found a valid schedule}
	\State $dp \gets append(dp, (C_i, C_j))$
    \State $dt \gets append(dt, a_i, a_j)$
    \State $tc \gets \phi$
\Else
    \State $tc \gets append(tc, c_t)$
\EndIf
\State $s' \gets \{dp, dt, tc\}$
\State $r_e \gets$ extrinsic reward from environment
\State Store transition ($s$, $c_t$, $r_e$, $s'$) in $R_M$

\State Sample minibatch of transitions from $R_M$ and update $Q_1$ weights 

\State $s \gets s'$
\State $t \gets t + 1$

\EndWhile

\EndFor

\end{algorithmic}
\end{algorithm}

All controllers share the same replay buffer and the same weights. Additionally, by randomly sampling meta-controller constraints, the controllers can be pre-trained to communicate and complete subtasks prior to the joint training as described above. In this case, $Q_2$ will start with these pre-trained weights rather than being initialized to be random in the above algorithm.

For the distributed scheduling task as described in the experiments, the $Critic$ provides an intrinsic reward $r_{intrinsic}=1.0$ only if (i) the controllers' actions are valid according to their constrained databases ($a_i \in D_i \wedge c_t$ and $a_j \in D_j \wedge c_t$), and (ii) the actions are in the correct order ($a_i < a_j$). For $NextControllerPair$, we use the heuristic of emitting controller pairs in the order expected by the environment: $\{(C_1, C_2), \dots, (C_{N-1}, C_N)\}$. Alternatively, a separate Q-network could be trained to select controller pairs, which would be useful in domains where a sequencing of controllers for pairwise communication is not manifest from the task description.

\subsection{Results}
\begin{figure}[t]
  \center
  \includegraphics[width=0.32\linewidth]{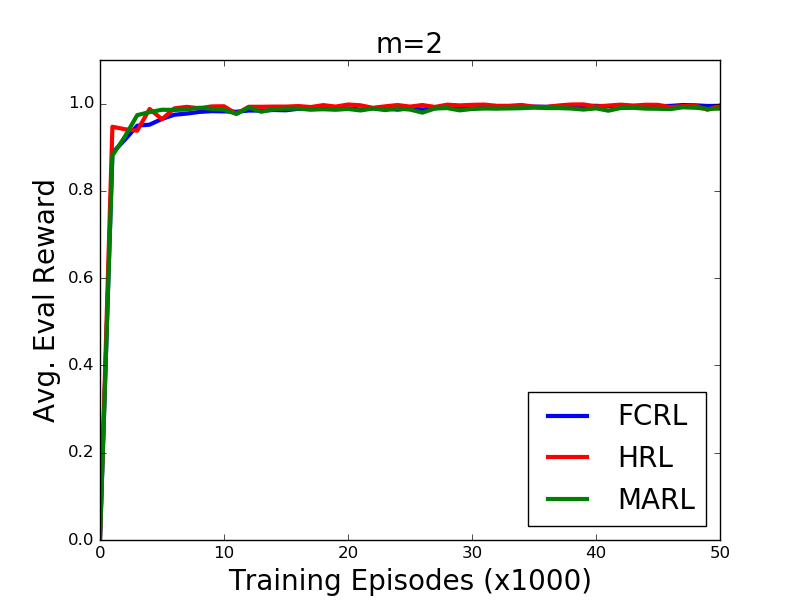}
  \includegraphics[width=0.32\linewidth]{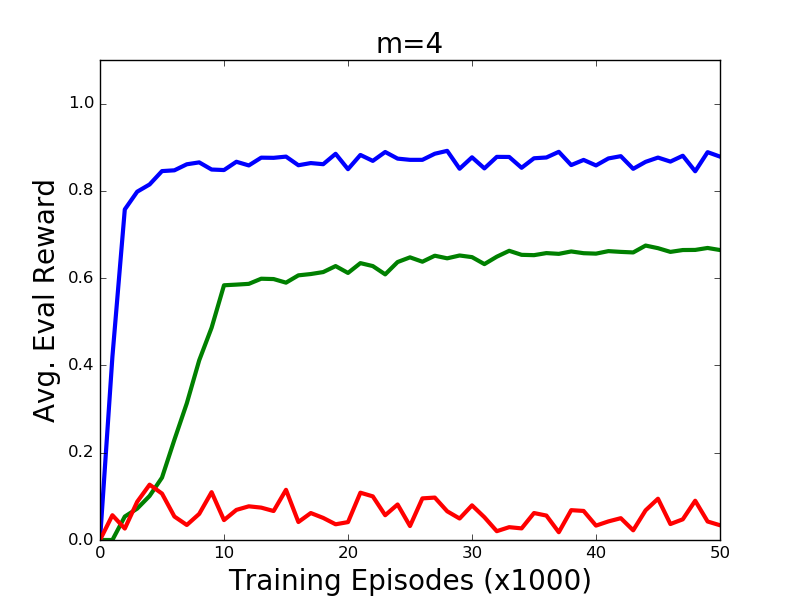}
  \includegraphics[width=0.32\linewidth]{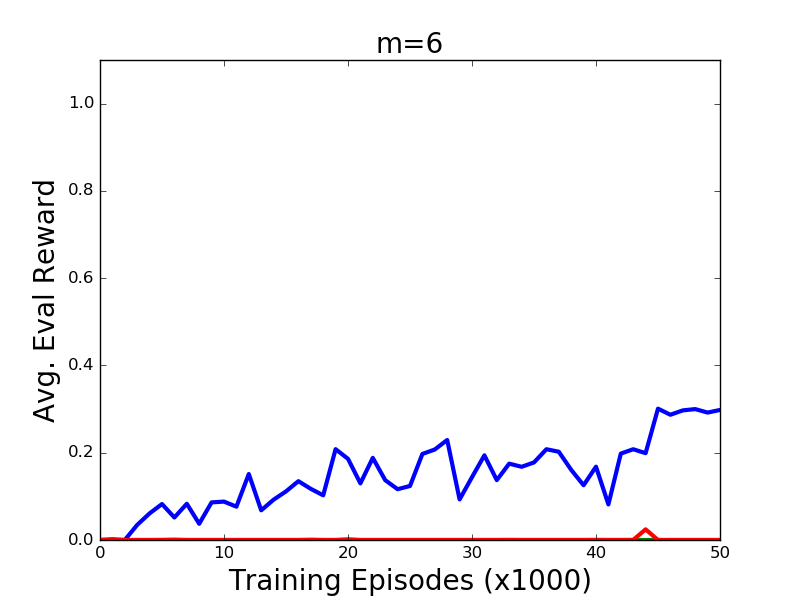}
  \caption{Comparing FCRL with baselines MARL and HRL, on three environments: (a) easy (m=2), (b) medium (m=4), and (c) hard (m=6).}
  \label{fig:results}
\end{figure}

We alternate training and evaluation for 1000 episodes each. Figure \ref{fig:results} plots the average reward on the evaluation episodes over the course of training. Each curve is the average of 5 independent runs with the same configuration. We ran three experiments by varying $m$, i.e. the number of agents that are part of the requested schedule and must communicate to come up with a valid plan.

For $m=2$, all three approaches are able to find the optimal policy as it requires only two agents to communicate. For $m=4$, HRL performs poorly as there is no inter-agent communication and the meta-controller must do all the work of picking the right sequence of constraints to surface a valid schedule. MARL does better as agents can communicate their preferences and get a chance to update their choices based on what other agents picked. FCRL does better than both baselines, as the meta-controller learns to guide the communications by constraining each pair of agents to focus on disjoint slices of the database, while the controllers have to only communicate with one other controller making it easy to agree upon a good pair of actions.

For $m=6$, both HRL and MARL are unable to see a positive reward, as finding a valid schedule requires significantly more exploration for the meta-controller and controller, respectively. FCRL is able to do better by dividing the problem into disjoint subtasks and leveraging temporal abstractions. However, the meta-controller's optimal policy is more intricate in this case, as it needs to learn to start with smaller constraint windows and try larger ones if the smaller one fails, so that the earlier agent pairs do not choose farther apart times when closer ones are possible.

\section{Discussion}
We presented a framework for combining hierarchical and multi-agent RL to benefit from temporal abstractions to reduce the communication complexity for finding globally consistent solutions with distributed policies. Our experimental results show that this approach scales better than baseline approaches as the number of communicating agents increases.


\textbf{Future work} The effect of increasing the size of the database or number of communication rounds will be interesting to study. Multi-agent training creates a non-stationary environment for the agent as other agents' policies change over the course of training. While we employ the standard DQN algorithm \cite{mnih2013dqn} to train the meta-controller, the controllers can be trained using recent policy gradient based methods (eg. counterfactual gradients \cite{foerster2017counterfactual} ) which address this problem.

\bibliography{nips_federated}
\bibliographystyle{abbrv}

\end{document}